\pdfoutput=1

\documentclass[11pt]{article}

\usepackage[final]{latex/acl}

\usepackage{booktabs}
\usepackage{amsmath}
\usepackage{times}
\usepackage{latexsym}
\usepackage{booktabs}
\usepackage{multirow}
\usepackage{algorithm}
\usepackage{algpseudocode}

\usepackage{xcolor}
\usepackage{tikz}

\usetikzlibrary{positioning, fit, backgrounds, decorations.pathreplacing, shadows, calc}

\usepackage[utf8]{inputenc}
\usepackage{microtype}
\usepackage{inconsolata}
\usepackage{graphicx}
\usepackage{authblk}

\title{Unsupervised Style Representation Learning \\ for AI-Text Detection via Paraphrase Inversion}

\author[1,2]{\bf Rafael Rivera Soto}
\author[1]{\bf Barry Chen}
\author[2]{\bf Nicholas Andrews}
\affil[1]{Lawrence Livermore National Laboratory}
\affil[2]{Johns Hopkins University}
\affil[]{\texttt{rafaelriverasoto@jhu.edu, chen52@llnl.gov, noa@jhu.edu}}

\begin{document}
\maketitle
\begin{abstract}
The rapid development of large language models (LLMs) has raised concerns about misuse such as plagiarism, misinformation, and automated influence operations, motivating the need for robust detectors.
Recent work has shown that neural representations of writing style are effective for detection and, crucially, robust to adversarial attacks that defeat most existing detectors.
However, current style-based detectors rely on authorship labels for training, and are limited to few-shot inference for detection, requiring in-distribution samples that may not always be available.
We learn discriminative style features without authorship labels by training a style encoder to reconstruct human-authored text from its machine-generated paraphrase; freezing a semantic encoder during training biases the style encoder to capture only the non-semantic features needed for reconstruction.
We evaluate the learned representations via two detection strategies: a few-shot detector and a zero-shot DeepSVDD-based detector.
Across benchmarks, our method matches or outperforms all baselines in the few-shot setting and, in the zero-shot regime, is competitive with fully supervised classifiers on in-distribution test data while generalizing better to unseen LLMs.
Beyond detection, the learned representations generalize to unseen tasks, achieving competitive performance on authorship verification and fine-grained style discrimination despite never being trained on either objective.\footnote{Code available at \url{https://github.com/rrivera1849/lusr}}
\end{abstract}

\section{Introduction}

The rapid advancement of large language models has raised concerns about their misuse, including academic plagiarism, the generation of misinformation~\citep{chen2024llmmisinformation}, and automated influence operations~\citep{goldstein2023generative}.
To counter these threats, a variety of machine-text detection methods have been proposed~\citep{wu2025survey}.
Zero-shot statistical methods, such as Binoculars~\citep{hans2024spotting} and FastDetectGPT~\citep{bao2024fastdetectgptefficientzeroshotdetection}, rely on the predictive distributions of LLMs, while supervised classifiers train on labeled human and machine text.
However, both families are brittle to simple adversarial attacks such as paraphrasing~\citep{krishna2020reformulatingunsupervisedstyletransfer, sadasivan2025aigeneratedtextreliablydetected}.
Supervised classifiers are further limited by sensitivity to the unavoidable distribution shifts that arise from the continual release of new LLMs, domains, and topics.
These fragilities suggest that current approaches rely on surface-level artifacts that are easily removed or that fail to generalize, motivating the need for detection methods grounded in more persistent features.

To this end,~\citet{soto2024fewshot} shifted the detection paradigm to a few-shot setting, showing that neural representations of \emph{writing style}, trained on the task of authorship verification, can detect machine-generated text given as few as five in-distribution machine-generated samples.
Crucially, these stylistic features appear robust, persisting even when models are prompted to adopt specific personas or when text is paraphrased.
However, their style representations require authorship labels for training, and detection cannot operate in zero-shot scenarios where no reference data for the target generator exists.
In this paper, we address both shortcomings: we propose a label-free training objective for learning style representations, and we introduce a DeepSVDD-based~\citep{ruff2018deep} detector that extends style-based detection to the zero-shot regime while retaining few-shot detection when support examples are available.

In this work we define style as those features which a paraphrase changes.
A semantically equivalent paraphrase preserves the meaning of a text but alters its surface form (punctuation, syntax, lexical choice, and others), which are precisely the features that distinguish one author's writing from another.
Therefore, an encoder that learns to recover what a semantically equivalent paraphrase changed must encode style.
Our approach, Unsupervised Style Representation Learning (USRL, which we permute to the more pleasant LUSR), operationalizes this idea via a paraphrase inversion task: given a machine-generated paraphrase of a human-authored text, the model reconstructs the original conditioned on semantic and stylistic representations of the original.
By freezing the parameters of the semantic encoder during training, we force the style encoder to capture only the non-semantic features needed for reconstruction, achieving a separation of style and content without authorship labels.

We evaluate LUSR in two detection settings.
First, using the few-shot detector of~\citet{soto2024fewshot}, we assess the discriminative power of the LUSR embeddings without further finetuning.
Second, following~\citet{zeng2025human}, we introduce a zero-shot detector based on DeepSVDD~\citep{ruff2018deep} that learns a compact hypersphere around machine-generated text in the LUSR embedding space, marking larger deviations from the hypersphere as human-authored.
Across the M4 and MAGE benchmarks, LUSR is the strongest few-shot detector among the style representations we evaluate and is competitive with a fully supervised classifier in the zero-shot regime while generalizing notably better than that classifier to unseen LLMs.
In~\autoref{sec:analysis}, we demonstrate that without any task-specific supervision, LUSR is competitive with representations trained for authorship verification and fine-grained style discrimination.
Taken together, this argues that paraphrase inversion as a task helps learn general stylistic features that are  useful over multiple tasks.

\begin{figure*}[t]
\centering
\definecolor{lusrcolor}{HTML}{8E5A9B}   
\definecolor{semcolor}{HTML}{34495E}    
\definecolor{lmcolor}{HTML}{2E86C1}     
\definecolor{plaincolor}{HTML}{ECF0F1}  
\definecolor{arrowcolor}{HTML}{566573}  
\begin{tikzpicture}[
    >=stealth,
    font=\small,
    box/.style={
      rounded corners=3pt, align=center,
      minimum height=10mm, minimum width=30mm,
      drop shadow={shadow xshift=0.5mm, shadow yshift=-0.5mm, fill=black!50, opacity=0.2}
    },
    lusrbox/.style={box, fill=lusrcolor, text=white, font=\small\bfseries},
    sembox/.style={box, fill=semcolor, text=white, font=\small\bfseries},
    lmbox/.style={box, fill=lmcolor, text=white, font=\small\bfseries, minimum height=16mm},
    plain/.style={rounded corners=3pt, align=center, draw=semcolor!60, fill=plaincolor,
                  minimum height=10mm, minimum width=28mm, font=\small\bfseries, text=semcolor},
    arr/.style={->, semithick, arrowcolor, rounded corners=4pt}
  ]

  \node[lusrbox]                   (lusr) {LUSR\\Representation};
  \node[sembox,  below=4mm of lusr] (sem)  {Semantic Representation\\{\scriptsize(Frozen)}};
  \node[plain,   below=4mm of sem]  (para) {Paraphrase(Text)};

  \node[lmbox, right=16mm of sem]   (clm)  {Causal\\Language Model};

  \node[plain, right=12mm of clm]   (out)  {Text};

  \draw[arr] (lusr.east) -- ([yshift=5mm]clm.west);
  \draw[arr] (sem.east)  -- (clm.west);
  \draw[arr] (para.east) -- ([yshift=-5mm]clm.west);

  \draw[arr] (clm.east) -- (out.west);

  \coordinate (top)     at ($(lusr.north)+(0,6mm)$);
  \coordinate (leftcol) at ($(lusr.west)+(-8mm,0)$);
  \coordinate (lusrL)   at (leftcol |- lusr);
  \coordinate (semL)    at (leftcol |- sem);

  \draw[semithick, arrowcolor, rounded corners=4pt]
    (out.north) -- (out.north |- top) -- (top) -- (leftcol |- top) -- (semL);
  \draw[arr] (lusrL) -- (lusr.west);
  \draw[arr] (semL)  -- (sem.west);
\end{tikzpicture}
\caption{
\textbf{Schematic of the paraphrase inversion training objective}: The model is trained to reconstruct the original human text (right) given its machine-generated paraphrase and two latent representations derived from the original text. 
By freezing the parameters of the semantic representation, the architecture introduces an inductive bias that helps the learnable LUSR representation encode only the residual, non-semantic features necessary to revert the machine paraphrase back to its human original.
}
\label{fig:concept}
\end{figure*}
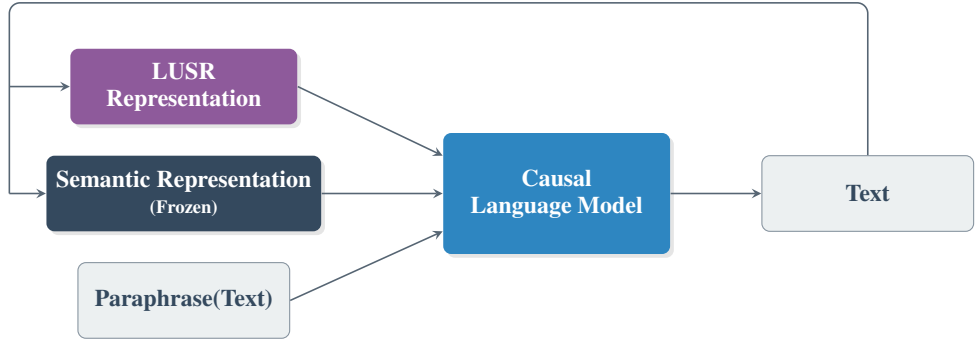

\section{Related Work}

\paragraph{Machine-Text Detection}
Existing approaches to detecting machine-generated text broadly fall into three categories.
\emph{Zero-shot statistical methods} exploit properties of a generating model's output distribution without any task-specific training.
Early work used token-level log-probabilities and rank statistics~\citep{solaiman2019releasestrategiessocialimpacts, ippolito-etal-2020-automatic, gehrmann2019gltrstatisticaldetectionvisualization}, while more recent methods leverage probability curvature~\citep{bao2024fastdetectgptefficientzeroshotdetection} or cross-model perplexity comparisons~\citep{hans2024spotting}.
These methods require access to a reference LLM at inference time and degrade when the true model is unknown or when text has been edited or paraphrased~\citep{sadasivan2025aigeneratedtextreliablydetected}.
\emph{Supervised classifiers} fine-tune a language model on labeled human and machine text~\citep{li-etal-2024-mage, hu2023radar, lee2024remodetect}, achieving strong in-distribution performance but suffering from distribution shifts across generators, domains, and adversarial attacks~\citep{NEURIPS2024_b61bdf7e}.
A third line of work uses \emph{watermarking} to embed detectable signals during generation~\citep{kirchenbauer2024watermarklargelanguagemodels, kuditipudi2024robustdistortionfreewatermarkslanguage}, but this requires control over the generation process and cannot be applied retroactively.
Orthogonal to all three categories, \citet{soto2024fewshot} demonstrated that neural representations of writing style can detect machine text in a few-shot setting, with robustness to paraphrasing and persona prompting.

\paragraph{Style Representation Learning}
Prior work in style representation learning typically follows a metric learning paradigm, utilizing contrastive objectives~\citep{oord2019representationlearningcontrastivepredictive, NEURIPS2020_d89a66c7} to enforce that embeddings of documents by the same author are clustered together, while those of different authors are pushed apart~\citep{rivera-soto-etal-2021-learning, wegmann-etal-2022-author, kim-etal-2025-leveraging, STAR}.
However, a well-documented problem with this approach is the entanglement of style with semantic content (e.g., where the topic itself is predictive of the author).
To mitigate this, recent studies have resorted to hard-positive and hard-negative mining strategies, attempting to construct training pairs where style is the sole discriminative variable~\citep{wegmann-etal-2022-author, fincke2024separatingstylesubstanceenhancing, liu-etal-2025-counterfactual}.
All of these methods require authorship labels for training.
Instead, LUSR learns style features through a paraphrase inversion objective.

\section{Unsupervised Style Representation Learning}\label{sec:usrl}

Our goal is to learn a style representation that is discriminative for machine-text detection without requiring authorship labels or contrastive sampling for representation learning.
We first describe the training objective used to learn the LUSR encoder (\S\ref{sec:usrl_train}), then present a few-shot detector (\S\ref{sec:usrl_detect_fewshot}) and a zero-shot DeepSVDD-based detector (\S\ref{sec:usrl_detect_zeroshot}) that operate on the learned representations.

\subsection{Training Objective}\label{sec:usrl_train}
\citet{rivera-soto-etal-2025-mitigating} explored paraphrase inversion---training an LLM to recover human-authored text from its machine-generated paraphrase---as a means of mitigating paraphrase attacks on machine-text detectors.
We repurpose this task as a training objective for learning style representations by augmenting it with two auxiliary encoders (see~\autoref{fig:concept}).
The intuition is that paraphrases preserve the meaning of text while altering its style.
By training a style encoder to help recover what the paraphrase changes, we target the stylistic features of the original text without requiring authorship labels.

\paragraph{Inputs and Architecture}
Given a human-authored text $t$ and its machine-generated paraphrase $P(t)$, we train a causal language model $g_{\psi}$ to reconstruct $t$ conditioned on $P(t)$, a stylistic embedding $\mathbf{e}_s = \text{LUSR}(t)$, and a semantic embedding $\mathbf{e}_l = \text{Sem}(t)$.
The LUSR encoder is initialized from \texttt{RoBERTa-large} and produces $\mathbf{e}_s$ via mean pooling over its token representations.
The semantic encoder $\text{Sem}(\cdot)$ is a frozen pre-trained sentence encoder (SBERT~\citep{reimers-gurevych-2019-sentence}).\footnote{We use \texttt{Gameselo/STS-multilingual-mpnet-base-v2} as our SBERT model.}
Both embeddings are projected through learned linear maps $\mathbf{W}_s$ and $\mathbf{W}_l$ to the hidden dimensionality of the causal LM and prepended to the token embeddings of the paraphrase:
\begin{equation}
\mathbf{X} = \bigl[\, \mathbf{W}_s \mathbf{e}_s \,;\, \mathbf{W}_l \mathbf{e}_l \,;\, \mathbf{E}(P(t)) \,\bigr],
\end{equation}
where $\mathbf{E}(\cdot)$ denotes the causal LM's token embedding lookup and $[\cdot\,;\,\cdot]$ denotes sequence-wise concatenation.

\paragraph{Reconstruction Objective}
The causal LM is trained with the standard teacher-forced cross-entropy loss to maximize the likelihood of the original text given its input $\mathbf{X}$.
We use Llama-3.2-1B~\citep{grattafiori2024llama3} as the causal LM, fine-tuned with LoRA~\citep{hu2022lora}.\footnote{For LoRA, we use $r=32$ and $\alpha=64$.}
Only the semantic encoder is fully frozen during training.
The LUSR encoder, both projection maps $\mathbf{W}_s$ and $\mathbf{W}_l$, and the LoRA adapters of the causal LM are jointly updated to minimize the reconstruction loss.
Freezing the semantic encoder forces the LUSR encoder to learn those features which are \emph{not} encoded in the semantic representation but are useful for reconstruction, that is, LUSR is biased to encode stylistic features.

\subsection{Few-Shot Detection}\label{sec:usrl_detect_fewshot}
To assess the discriminative power of the LUSR embedding space without additional learned parameters, we follow the few-shot approach of~\citet{soto2024fewshot} (see App.~\autoref{fig:detection_protocol} for an illustration).
Given a limited number $k$ of in-distribution AI-generated support samples ($k \in [1, 5]$), we employ a prototype-based approach~\citep{10.5555/3294996.3295163}.
We compute the centroid of the support samples in LUSR space and use cosine similarity to this centroid as the detection score.
When multiple support sets are available (e.g., when discriminating among multiple LLMs), we compute a separate prototype for each class and assign the detection score as the maximum similarity to any prototype.

\subsection{Zero-Shot Detection via DeepSVDD}\label{sec:usrl_detect_zeroshot}
Since humans exhibit more diverse writing styles than LLMs, machine-generated text tends to cluster in style space while human text is dispersed~\citep{reinhart2025, wang-etal-2025-catch}.
This motivates an out-of-distribution (OOD) detection framing, in which machine-generated text is treated as in-distribution (ID) and human-authored text as OOD.
We adopt the OOD detection framework of~\citet{zeng2025human}, replacing their feature extractor with the LUSR encoder. 
Specifically, we attach a learned projection head $\phi$ on top of the LUSR encoder and fine-tune the entire LUSR encoder above its embedding layer together with $\phi$.
The center $c$ is initialized once from the projected embeddings of machine-generated training samples produced by the pretrained LUSR encoder and held fixed thereafter.
The network is then trained with the equally weighted combined objective $\mathcal{L} = \mathcal{L}_{\text{DSVDD}} + \mathcal{L}_{\text{InfoNCE}}$ until convergence, where the Deep Support Vector Data Description (DeepSVDD)~\citep{ruff2018deep} loss pushes all machine-generated samples to the center $c$:
\begin{equation}
\mathcal{L}_{\text{DSVDD}} = \frac{1}{N}\sum_{i=1}^{N} \| \phi(\text{LUSR}(x_i)) - c \|^2,
\end{equation}
where $x_i$ is the $i$-th machine-generated training sample, and $\mathcal{L}_{\text{InfoNCE}}$~\citep{oord2019representationlearningcontrastivepredictive} treats machine-generated samples as positives and human-authored samples as negatives within each batch, encouraging separation between the two classes in the projected space.
At inference, we compute the distance of a test sample's projected embedding to $c$.
Samples far from the center are flagged as out-of-distribution and thus classified as human-authored.

\section{Experimental Details}\label{sec:experimental_details}

\paragraph{Training Dataset: LUSR} 
To train LUSR, we use a subset of the Reddit Million User Dataset (MUD)~\citep{khan-etal-2021-deep}, an established corpus for training authorship representations.
We retain only posts containing between $32$ and $128$ tokens, as determined by the \texttt{roberta-large} tokenizer.
From the filtered pool, we randomly sample $16$ posts per author.
To ensure a diverse representation of writing styles, we perform stratified sampling across authorship clusters to arrive at a final training set comprising $63,184$ unique authors (see~\autoref{sec:app_data_curation} for details on the clustering procedure).
To generate paraphrases, we prompt \texttt{Mistral-7B-Instruct} to paraphrase each sample\footnote{We set the top-p to $0.9$ and the temperature to $0.7$, generating $5$ paraphrases per sample.}.
We ablate the choice of paraphraser in~\autoref{sec:ablations}.

\paragraph{Training Dataset: DSVDD Detector} To train the DeepSVDD detector introduced in~\autoref{sec:usrl_detect_zeroshot}, we utilize the training split of the MAGE dataset (described below).

\paragraph{Evaluation Datasets}
We evaluate on two benchmarks.
The MAGE~\citep{li-etal-2024-mage} dataset includes text generated by $27$ LLMs (e.g., GPT-4, LLaMA, Pythia) alongside human-written content across $10$ domains, totaling $332$K training and $57$K test samples.
It defines six evaluation scenarios including cross-domain, unseen-domain, and unseen-model settings, and others.
Following~\citet{soto2024fewshot}, we also use the M4~\citep{wang-etal-2023-m4} dataset augmented with Amazon reviews, covering $9$ LLMs across five domains.

\paragraph{Baselines for Few-Shot Evaluation (\autoref{sec:nonparametric})}
We follow~\citet{soto2024fewshot}'s few-shot evaluation protocol on M4 and adopt their full set of baselines.
The style-based detectors are CISR~\citep{wegmann-etal-2022-author} and three LUAR variants released by~\citet{soto2024fewshot}: LUAR CRUD, LUAR Multi-LLM, and LUAR Multidomain, along with ProtoNet~\citep{10.5555/3294996.3295163}, and SBERT~\citep{reimers-gurevych-2019-sentence}.
We additionally report results for the zero-shot and supervised baselines described below.

\paragraph{Baselines for Zero-Shot Evaluation (\autoref{sec:parametric})}
On MAGE, we compare against prominent zero-shot statistical detectors: Entropy~\citep{ippolito-etal-2020-automatic}, Rank~\citep{gehrmann2019gltrstatisticaldetectionvisualization}, Log-Rank~\citep{solaiman2019releasestrategiessocialimpacts}, LRR~\citep{su2023detectllm}, NPR~\citep{su2023detectllm}, Fast-DetectGPT~\citep{bao2024fastdetectgptefficientzeroshotdetection}, Binoculars~\citep{hans2024spotting}, DNA-GPT~\citep{yang2024dnagpt}, and Revise-Detect~\citep{zhu-etal-2023-beat}.
We also compare against supervised fine-tuned classifiers: the Longformer classifier of~\citet{li-etal-2024-mage}, RADAR~\citep{hu2023radar}, and ReMoDetect~\citep{lee2024remodetect}.
Furthermore, motivated by the finding that style representations are discriminative of machine-generated text out-of-the-box~\citep{soto2024fewshot}, we additionally evaluate $k$-nearest neighbor (KNN) classifiers trained on the MAGE training set over LUAR~\citep{rivera-soto-etal-2021-learning}, MSR~\citep{kim-etal-2025-leveraging}, CISR~\citep{wegmann-etal-2022-author}, StyleDistance~\citep{patel-etal-2025-styledistance}, and LUSR embeddings, sweeping over $k$ on the validation set (see~\autoref{sec:app_knn} for details).

\paragraph{Metrics} To evaluate the performance of machine-text detectors, we use the receiver operating characteristic curve (ROC), which summarizes the trade-off between the false positive rate (FPR) and true positive rate (TPR) at different  operating thresholds.
To summarize this curve, we calculate the area under the curve restricted to the range of thresholds that don't exceed 1\% FPR, which we denote by AUROC(1).
This metric captures the performance in real-world settings where false-positives are costly, and has become a standard metric in the relevant literature~\citep{hans2024spotting, soto2024fewshot, dugan-etal-2024-raid}.
When calculating the full area (across all FPRs), we refer to it as AUROC.

\section{Few-Shot Evaluation of LUSR Embeddings}\label{sec:nonparametric}

Before introducing any learned parameters, we evaluate the discriminative power of the LUSR embedding space using the few-shot detection protocol proposed by~\citet{soto2024fewshot} on the M4 dataset described in~\autoref{sec:experimental_details}.
This isolates the quality of the representations from the capacity of any downstream model.
Crucially, the style representations are trained exclusively on Reddit data, meaning that \emph{all evaluation domains and all LLMs are unseen during representation learning}.
We evaluate in three settings that reflect realistic deployment scenarios.
All few-shot baselines in this section use the same prototype-based scoring described in~\autoref{sec:usrl_detect_fewshot}, differing only in the underlying style embedding.

\paragraph{Single-Target Detection}
In this setting, we assume access to a small support set of $k$ examples known to originate from a specific LLM of concern (e.g., ChatGPT).
The objective is to identify further samples generated by this same model from among a larger collection.
For each evaluation domain and each LLM contributing to that domain, we perform $1000$ random trials.
In each trial, we randomly sample $k$ samples from that LLM to serve as the support set, with all remaining samples from that LLM and all human-authored samples serving as \emph{queries}.

\paragraph{Multiple-Target Detection}
In this setting, the goal is to detect text as having originated from \emph{any} of multiple LLMs simultaneously.
For each LLM contributing to that domain, we randomly select a support set of size $k$, leaving all other samples as queries.
We perform $1000$ random trials.

\paragraph{Robustness to Paraphrasing}
This setting is identical to the single-target setting, except that the queries may have been paraphrased.
We paraphrase a proportion (e.g. $20\%$ or $80\%$) of the queries using DIPPER\footnote{We set DIPPER's hyperparameters to $\text{L}=20$, and $\text{O}=0$.}, a paraphrasing model originally designed to fool detectors.
We evaluate two cases: one where the support set consists of the original LLM outputs, and another where two support sets are given, one consisting of original LLM outputs and another consisting of DIPPER paraphrases.
We perform $1000$ random trials.

\begin{table}[t!]
\centering
\caption{\textbf{Single-target few-shot detection.}
AUROC(1) is reported for zero-shot and supervised baselines (no support set) and for few-shot methods with support sizes $k \in \{1, 5\}$.
All few-shot evaluations are on domains and LLMs unseen during style representation training.}
\label{tab:single_target}
\begin{tabular}{lcc}
\toprule
\textbf{Model} & \multicolumn{2}{c}{\textbf{AUROC(1)}} \\
\midrule
Binoculars & \multicolumn{2}{c}{\bf 69} \\
FastDetectGPT & \multicolumn{2}{c}{65} \\
Rank & \multicolumn{2}{c}{50} \\
LogRank & \multicolumn{2}{c}{50} \\
LRR & \multicolumn{2}{c}{50} \\
Revise-Detect & \multicolumn{2}{c}{60} \\
DNA-GPT & \multicolumn{2}{c}{51} \\
\midrule
\textit{Supervised Classifiers}
Rank & \multicolumn{2}{c}{50} \\
Longformer & \multicolumn{2}{c}{58} \\
RADAR & \multicolumn{2}{c}{50} \\
RemoDetect & \multicolumn{2}{c}{64} \\
\midrule
\textbf{Support Size $\rightarrow$} & \textbf{1} & \textbf{5} \\
\midrule
LUAR CRUD & 60 & 87 \\
LUAR Multi-LLM & 61 & 88 \\
LUAR Multidomain & 60 & 89 \\
CISR & 58 & 84 \\
ProtoNet & 61 & 87 \\
SBERT & 52 & 62 \\
\textbf{LUSR} & \textbf{69} & \textbf{96} \\
\bottomrule
\end{tabular}
\end{table}

\subsection{Single-Target Results}\label{sec:nonparametric_single}

\autoref{tab:single_target} presents the single-target detection results.
Several findings emerge from these results.
First, LUSR achieves an AUROC(1) of $69$ with only a single support sample, outperforming all baselines except Binoculars, whose score it matches.
This is notable because, with $k=1$, the detector operates from a single reference sample yet already matches methods such as Binoculars ($69$) and Fast-DetectGPT ($65$) that require access to an LLMs probability distribution.
The gap widens substantially at $k=5$ where LUSR achieves $96$, a $7$-point improvement over the best prior style-based method (LUAR Multidomain at $89$).

\begin{table}[t!]
\centering
\caption{\textbf{Multiple-target few-shot detection.}
AUROC(1) is reported with support sizes $k \in \{1, 5\}$.
LUSR maintains a substantial lead over all prior style representations in this more challenging multi-generator setting.}
\label{tab:multi_target}
\begin{tabular}{lcc}
\toprule
\multirow{2}{*}{\textbf{Model}} & \multicolumn{2}{c}{\textbf{AUROC(1)}} \\
\cmidrule(lr){2-3}
 & \textbf{1} & \textbf{5} \\
\midrule
LUAR CRUD        & 60 & 82 \\
LUAR Multi-LLM   & 58 & 82 \\
LUAR Multidomain & 60 & 83 \\
CISR             & 58 & 81 \\
SBERT            & 50 & 57 \\
ProtoNet         & 55 & 64 \\
\textbf{LUSR}    & \textbf{67} & \textbf{93} \\
\bottomrule
\end{tabular}
\end{table}

\subsection{Multiple-Target Results}\label{sec:nonparametric_multi}

\autoref{tab:multi_target} presents results for the multiple-target setting.
Note that zero-shot statistical methods and supervised classifiers are not directly comparable in this setting, as they do not possess a mechanism adaptively select which LLMs to detect.
Only embedding-based methods, which construct per-LLM prototypes can operate in this setting.
LUSR continues to outperform all competing representations by a wide margin.
At $k=5$, LUSR achieves an AUROC(1) of $93$, a $10$-point improvement over the best prior style representation ($83$ with LUAR Multidomain).
Even at $k=1$, LUSR ($67$) surpasses all other methods at the same support size.

\subsection{Robustness to Paraphrase Attacks}\label{sec:nonparametric_paraphrase}

\begin{figure}[t!]
\centering
\includegraphics[width=\columnwidth]{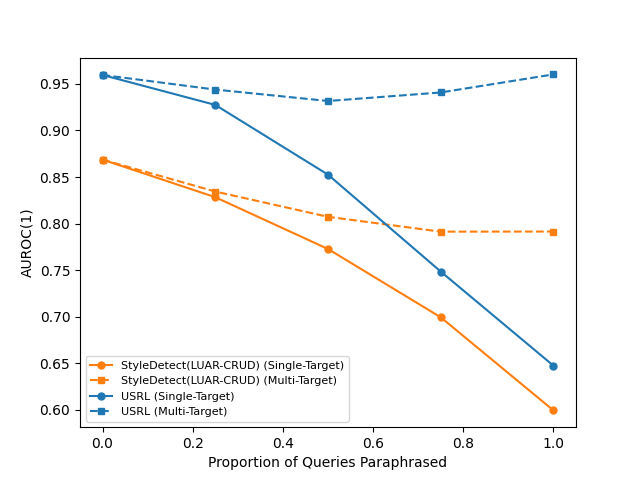}
\caption{\textbf{Robustness to paraphrase attacks.}
Mean AUROC(1) as the proportion of paraphrased query episodes increases.
The multiple-target variants, which include paraphrased LLM text in the support set, are substantially more robust.
LUSR consistently outperforms LUAR CRUD and ProtoNet across all paraphrase proportions.}
\label{fig:paraphrase_robustness}
\end{figure}

\autoref{fig:paraphrase_robustness} presents the results.
As in the multiple-target setting, only embedding-based methods are applicable here (see~\autoref{sec:nonparametric_multi}).
As expected, all methods experience a decline in performance as the proportion of paraphrased queries increases, since paraphrasing removes some of the surface-level stylistic cues that distinguish machine from human text.
However, two key observations emerge.
First, LUSR consistently outperforms LUAR CRUD at every paraphrase proportion in both single-target and multiple-target settings.
Second, the multiple-target variants exhibit substantially greater robustness than their single-target counterparts.
This is because the multiple-target formulation incorporates paraphrased LLM text in the support set, allowing the detector to match queries against both original and paraphrased stylistic signatures.
Notably, LUSR in the multiple-target setting maintains an AUROC(1) above $90$ even when the majority of queries have been paraphrased.

Across all three settings, LUSR consistently outperforms all other style representations without any additional learned classification parameters and on evaluation domains and LLMs unseen during training, indicating that the paraphrase inversion objective produces an embedding space in which human and machine text are inherently well-separated.

\section{Zero-Shot Detection with DeepSVDD}\label{sec:parametric}

Having established that the LUSR embedding space is inherently discriminative in the few-shot setting, we now evaluate whether a DeepSVDD-based detector can further leverage these representations for zero-shot detection.
As stated in~\autoref{sec:experimental_details}, we train the DeepSVDD detector on the MAGE training set, and evaluate on the MAGE test set.
Note that the Longformer classifier and the KNNs are also trained on the same training set.

\paragraph{Main Results}

\autoref{tab:mage_results} presents results on the MAGE test set.
We compare the LUSR-based DeepSVDD detector against all zero-shot, supervised, and few-shot baselines described in~\autoref{sec:experimental_details}.

\begin{table}[t!]
\centering
\caption{\textbf{Results on the MAGE benchmark.}
AUROC(1) for zero-shot and supervised baselines (no support set) and for the LUSR DeepSVDD detector.
The LUSR DeepSVDD detector is competitive with the fully supervised Longformer classifier on in-distribution test data and, as shown in~\autoref{tab:generalization}, generalizes better to unseen LLMs.}
\label{tab:mage_results}
\begin{tabular}{lc}
\toprule
\textbf{Model} & \textbf{AUROC(1)} \\
\midrule
\multicolumn{2}{l}{\emph{Zero-shot baselines}} \\
Binoculars & 54 \\
FastDetectGPT & 63 \\
Rank & 50 \\
LogRank & 50 \\
LRR & 50 \\
Revise-Detect & 50 \\
DNA-GPT & 51 \\
\midrule
\multicolumn{2}{l}{\emph{Supervised classifiers}} \\
Longformer & \bf 81 \\
RADAR & 50 \\
ReMoDetect & 53 \\
\midrule
\multicolumn{2}{l}{\emph{KNN style-based detectors}} \\
KNN (LUAR) & 65 \\
KNN (MSR) & 60 \\
KNN (CISR) & 58 \\
KNN (StyleDistance) & 66 \\
KNN (LUSR) & 73 \\
\midrule
\multicolumn{2}{l}{\emph{Ours}} \\
LUSR DeepSVDD & \underline{78} \\
\bottomrule
\end{tabular}
\end{table}

The LUSR DeepSVDD detector achieves an AUROC(1) of $78$, outperforming all zero-shot baselines and trailing only the fully supervised Longformer classifier ($81$).
As we show in~\autoref{tab:generalization}, the DeepSVDD detector also generalizes more readily than the MAGE classifier to unseen LLMs, indicating that style-based features capture generator-invariant cues rather than artifacts of any particular model.

\paragraph{Unseen Domains and Unseen Models}

To evaluate how well the DeepSVDD detector generalizes beyond its training distribution, we compare it against the Longformer supervised classifier on two of MAGE's held-out evaluation scenarios: unseen domains and unseen models.
\autoref{tab:generalization} reports the results.

\begin{table}[t!]
\centering
\caption{\textbf{Generalization to unseen domains and unseen models.}
AUROC on two MAGE evaluation scenarios.
We report full AUROC rather than AUROC(1) to enable direct comparison with results reported by~\citet{li-etal-2024-mage}, who do not provide AUROC(1).}
\label{tab:generalization}
\begin{tabular}{lc}
\toprule
\textbf{Model} & \textbf{AUROC} \\
\midrule
\textbf{Unseen Domains} \\
Longformer &  \bf 95 \\
LUSR DeepSVDD & 91 \\
\midrule
\textbf{Unseen Models} \\
Longformer & 93 \\
LUSR DeepSVDD & \bf 97 \\
\bottomrule
\end{tabular}
\end{table}

The two models exhibit complementary strengths.
Longformer performs better on unseen domains ($95$ vs.\ $91$), while the LUSR DeepSVDD detector outperforms Longformer on unseen models ($97$ vs.\ $93$), suggesting that style-based features generalize more readily to new generators.
This result is consistent with the hypothesis that writing style captures a fundamental distinction between human and machine text that is less dependent on the specific model used for generation.

\section{Does Paraphrase Inversion Capture Style More Broadly?}\label{sec:analysis}

\begin{table}[t!]
\centering
\footnotesize
\caption{\textbf{Unseen task evaluation.}
AUROC is the average score across all PAN13/14/15/20/21 authorship verification tasks.
STEL and S-o-C are averaged across various style axes.
Higher S-o-C indicates stronger content independence.
LUSR is never trained on authorship labels.}
\label{tab:authorship_stel}
\begin{tabular}{lccc}
\toprule
\textbf{Model} & \textbf{AUROC} & \textbf{STEL} & \textbf{S-o-C} \\
\midrule
\emph{Supervised} \\
LUAR & \bf 78 & 82 & 5 \\
MSR & 73 & 82 & 6 \\
CISR & 70 & 78 & 24 \\
StyleDistance & 73 & \bf 85 & \bf 30 \\
\midrule
\emph{Unsupervised} \\
LUSR & \underline{74} & \underline{84} & \underline{25} \\
\bottomrule
\end{tabular}
\end{table}

If paraphrase inversion teaches an encoder only to spot the surface artifacts of one paraphraser, the resulting representation is narrow.
If, instead, it captures \emph{style} in a more general sense, the representation should transfer to other tasks that depend on stylistic distinctions.
We therefore evaluate LUSR on two tasks unseen during training: authorship verification and fine-grained style discrimination.
A representation that handles both is one whose features are stylistic in a broad sense.

\paragraph{Authorship Verification}
We evaluate on the PAN13/14/15/20/21~\citep{pan2013,pan2014,pan2015,pan2020,pan2021} authorship verification tasks, where given two text samples the goal is to determine whether they were written by the same author.
Every representation we compare against \emph{except} LUSR is explicitly trained on authorship verification, so competitive performance from LUSR would indicate that the paraphrase inversion objective captures author-distinguishing features as a byproduct.
We report the average AUROC across all PAN tasks in~\autoref{tab:authorship_stel}.

LUSR achieves the second-highest AUROC ($74$), trailing only LUAR and outperforming MSR, CISR, and StyleDistance, despite never having seen a verification trial during training.
This is notable given the scale of supervision used by competing methods.
MSR and LUAR are trained on $4.5$ million and $1$ million authors, respectively.
These results suggest that the paraphrase inversion objective implicitly captures author-distinguishing features as a byproduct of learning to reconstruct style.

\paragraph{Style Feature Recognition (STEL)}
The STEL benchmark~\citep{wegmann-nguyen-2021-capture} evaluates whether embeddings capture fine-grained style features using curated natural text.
Given two anchor sentences and two test sentences, STEL measures whether the embedding correctly pairs each test sentence with the anchor that shares the same style, based on cosine similarity.
The harder STEL-or-Content (S-o-C) variant presents a single anchor alongside two test sentences: one that matches the anchor's style but differs in content, and another that is a paraphrase of the anchor (matching content but differing in style).
To succeed on S-o-C, a model must represent style features more strongly than content features~\citep{patel-etal-2025-styledistance}.
We report the STEL and S-o-C scores in~\autoref{tab:authorship_stel}.
LUSR is second-best on both STEL and S-o-C (\autoref{tab:authorship_stel}), trailing only StyleDistance which is explicitly trained for content independence via synthetic contrastive examples.

Across both tasks LUSR matches representations trained explicitly for them, indicating that paraphrase inversion captures \emph{general} stylistic features.

\section{Ablations}\label{sec:ablations}

We ablate two design decisions on the single-target few-shot detection task of~\autoref{sec:nonparametric_single}: (i) whether the semantic encoder is frozen during training, and (ii) the choice of paraphraser used to generate training pairs. \autoref{tab:ablations} reports both. 
Freezing the semantic encoder yields consistent improvements, validating its role as an inductive bias that pushes the style encoder toward non-semantic features. 
The choice of paraphraser, by contrast, only modestly affects performance.

\begin{table}[t!]
\centering
\footnotesize
\caption{\textbf{Ablations.} Single-target AUROC(1) for variants of LUSR: the semantic encoder is frozen vs.\ jointly trained, and the paraphraser used to generate training pairs is varied.}
\label{tab:ablations}
\begin{tabular}{lcc}
\toprule
\textbf{Variant} & $k=1$ & $k=5$ \\
\midrule
\emph{Semantic encoder} \\
\quad Frozen (default) & \bf 69 & \bf 96 \\
\quad Unfrozen        & 66 & 91 \\
\midrule
\emph{Paraphraser} \\
\quad Mistral-7B (default) & \bf 69 & \bf 96 \\
\quad Llama-3.2-3B        & 68 & \bf 96 \\
\quad Ministral-8B        & 68 & 93 \\
\bottomrule
\end{tabular}
\end{table}

\section{Conclusion}

We introduced LUSR, a training strategy that operationalizes style as \emph{what a paraphrase changes}: we train a style encoder to encode features that help recover a human-authored text from its machine-generated paraphrase.
This training scheme is notable in that it doesn't require authorship labels.
The resulting representations support both few-shot and zero-shot AI-text detection, and transfer well to authorship verification and fine-grained style discrimination tasks, suggesting that paraphrase inversion is a general-purpose, authorship-label-free signal for learning writing style.

\paragraph{Limitations}
Our style representations are trained exclusively on English-language Reddit data paraphrased by Mistral-7B-Instruct.
While the paraphraser ablation in~\autoref{sec:ablations} suggests only modest sensitivity to this choice, the training procedure still inherits the stylistic regularities of whichever LLM produces the paraphrases, and effectiveness on other languages, domains, and formal registers (e.g., legal or medical text) remains to be validated.
A further constraint of the paraphrase inversion objective is that what counts as ``non-semantic'' is determined by the choice of semantic encoder. Biases in SBERT's representation of meaning therefore propagate into what LUSR encodes as style.

\bibliography{latex/custom}
\newpage
\appendix

\section{Dataset Statistics}\label{sec:app_dataset_stats}
\autoref{tab:dataset_stats} reports the number of samples for each dataset.

\begin{table}[t!]
\centering
\footnotesize
\caption{\textbf{Dataset sample counts.} Number of samples used for LUSR training and for the few-shot (M4), zero-shot (MAGE), and unseen-task (PAN, STEL) evaluation splits.}
\label{tab:dataset_stats}
\begin{tabular}{llr}
\toprule
\textbf{Dataset} & \textbf{Split} & \textbf{\# Samples} \\
\midrule
Reddit (LUSR training) & Train & 1{,}010{,}944 \\
\midrule
M4   & Test  & 84{,}141 \\
\midrule
MAGE & Train & 319{,}071 \\
MAGE & Validation & 56{,}793 \\
MAGE & Test  & 56,820 \\
\midrule
PAN13/14/15/20/21 & Verification & 4,260 \\
STEL & Style Probes & 1,384 \\
\bottomrule
\end{tabular}
\end{table}

\section{Training Data Curation}\label{sec:app_data_curation}

To select a diverse subset of authors from the Reddit Million User Dataset, we embed each author using the LUAR\footnote{rrivera1849/LUAR-MUD} authorship representation and cluster the resulting embeddings with Affinity Propagation~\citep{Frey2007ClusteringBP}.
We then perform stratified sampling across clusters.
Note that LUAR is used solely for data selection and does not provide any training signal to LUSR --- the LUSR encoder never sees LUAR embeddings during training.
In principle, any method that groups authors by stylistic similarity could serve the same purpose.
We use LUAR for convenience as it is readily available.

\section{KNN Classifier Details}\label{sec:app_knn}

For each style embedding (LUAR, MSR, CISR, StyleDistance, LUSR), we train a $k$-nearest neighbor classifier on the MAGE training set and evaluate on the MAGE test set.
Embeddings are extracted via mean pooling over token representations.
We use cosine distance as the distance metric.
We sweep $k \in \{1, 2, 3, 5, 10, 25, 50, 100\}$ and select the value that maximizes AUROC(1) on the MAGE validation set.
The KNN classifiers are evaluated only on the MAGE benchmark; they do not appear in the M4 few-shot evaluation (\autoref{sec:nonparametric}), which uses the prototype-based scoring described in~\autoref{sec:usrl_detect_fewshot}.

\section{Few-Shot Detection Protocol}\label{sec:app_detection_protocol}

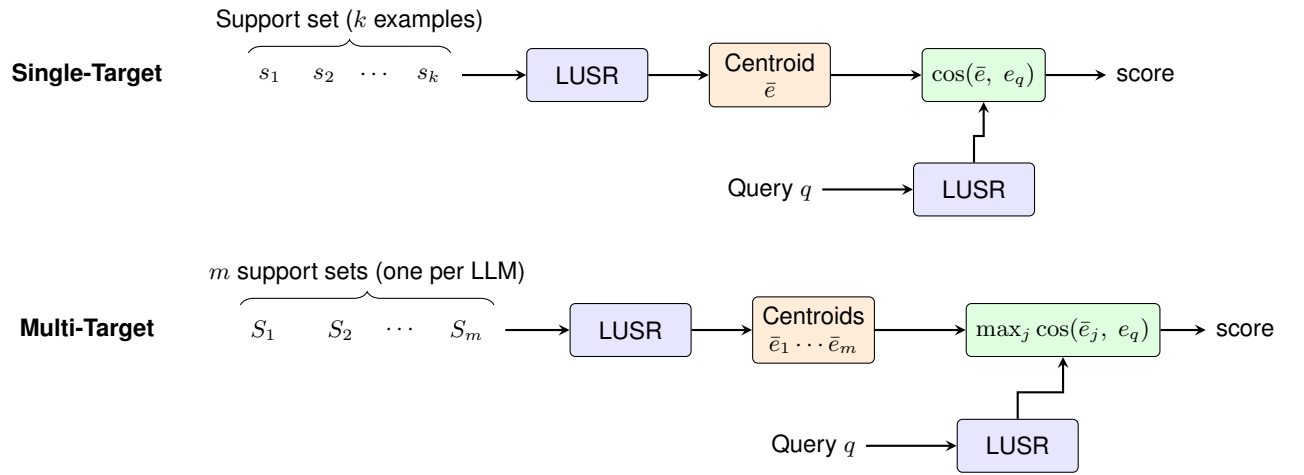
\begin{figure*}[ht]
\centering
\begin{tikzpicture}[
    node distance=0.6cm and 1.0cm,
    box/.style={draw, rounded corners=2pt, minimum height=0.7cm, minimum width=1.6cm, font=\small\sffamily, align=center},
    embed/.style={box, fill=blue!10},
    op/.style={box, fill=orange!15},
    score/.style={box, fill=green!12},
    doc/.style={font=\small\sffamily, align=center},
    arr/.style={->, >=stealth, thick},
    brace/.style={decorate, decoration={brace, amplitude=5pt, raise=2pt}},
]

\node[doc] (st_label) {\textbf{Single-Target}};

\node[doc, right=1.0cm of st_label] (s1) {$s_1$};
\node[doc, right=0.15cm of s1] (s2) {$s_2$};
\node[doc, right=0.05cm of s2] (sdots) {$\cdots$};
\node[doc, right=0.05cm of sdots] (sk) {$s_k$};

\draw[brace] (s1.north west) -- (sk.north east) node[midway, above=6pt, font=\small\sffamily] {Support set ($k$ examples)};

\node[embed, right=1.0cm of sk] (enc_s) {LUSR};

\node[op, right=0.8cm of enc_s] (centroid) {Centroid\\$\bar{e}$};

\node[score, right=1.2cm of centroid] (cossim) {$\cos(\bar{e},\; e_q)$};

\node[doc, right=0.8cm of cossim] (out_st) {score};

\node[doc, below=0.8cm of centroid] (q_st) {Query $q$};
\node[embed, right=1.2cm of q_st] (enc_q) {LUSR};

\draw[arr] (sk.east) ++(0.15,0) -- (enc_s.west);
\draw[arr] (enc_s) -- (centroid);
\draw[arr] (centroid) -- (cossim);
\draw[arr] (cossim) -- (out_st);
\draw[arr] (q_st) -- (enc_q);
\draw[arr] (enc_q.north) -- ++(0, 0.35) -| (cossim.south);

\node[doc, below=2.8cm of st_label] (mt_label) {\textbf{Multi-Target}};

\node[doc, right=1.0cm of mt_label] (set1) {$S_1$};
\node[doc, right=0.4cm of set1] (set2) {$S_2$};
\node[doc, right=0.15cm of set2] (setdots) {$\cdots$};
\node[doc, right=0.15cm of setdots] (setm) {$S_m$};

\draw[brace] (set1.north west) -- (setm.north east) node[midway, above=6pt, font=\small\sffamily] {$m$ support sets (one per LLM)};

\node[embed, right=1.0cm of setm] (enc_m) {LUSR};

\node[op, right=0.8cm of enc_m] (centroids) {Centroids\\$\bar{e}_1 \cdots \bar{e}_m$};

\node[score, right=1.2cm of centroids] (maxcos) {$\max_j \cos(\bar{e}_j,\; e_q)$};

\node[doc, right=0.6cm of maxcos] (out_mt) {score};

\node[doc, below=0.8cm of centroids] (q_mt) {Query $q$};
\node[embed, right=1.2cm of q_mt] (enc_q2) {LUSR};

\draw[arr] (setm.east) ++(0.15,0) -- (enc_m.west);
\draw[arr] (enc_m) -- (centroids);
\draw[arr] (centroids) -- (maxcos);
\draw[arr] (maxcos) -- (out_mt);
\draw[arr] (q_mt) -- (enc_q2);
\draw[arr] (enc_q2.north) -- ++(0, 0.35) -| (maxcos.south);

\end{tikzpicture}
\caption{
\textbf{Few-shot detection protocol.}
\emph{Top:} Single-target detection computes the centroid of $k$ support embeddings from one LLM and scores a query by cosine similarity to this centroid.
\emph{Bottom:} Multi-target detection maintains a separate centroid for each of $m$ LLMs and assigns the maximum similarity as the detection score.
}
\label{fig:detection_protocol}
\end{figure*}

\autoref{fig:detection_protocol} illustrates the few-shot detection protocol described in~\autoref{sec:usrl_detect_fewshot}.

\paragraph{Single-Target Detection}
Given $k$ support samples known to originate from a specific LLM, each sample is passed through the frozen LUSR encoder to obtain a style embedding.
The centroid $\bar{e}$ of these $k$ embeddings serves as the prototype for that LLM's writing style.
To score a query $q$, we encode it with the same LUSR encoder and compute the cosine similarity between the query embedding $e_q$ and the centroid $\bar{e}$.
A higher similarity indicates that the query is more likely to have been generated by the target LLM.

\paragraph{Multi-Target Detection}
When the goal is to detect text from any of $m$ LLMs simultaneously, we construct a separate support set $S_j$ for each LLM $j \in \{1, \ldots, m\}$ and compute a centroid $\bar{e}_j$ for each.
To score a query, we compute its cosine similarity to every centroid and take the maximum, $\max_j \cos(\bar{e}_j, e_q)$, as the detection score.
This allows the detector to flag text from any of the $m$ target LLMs without needing to know which specific model produced it.

\section{Use of AI Assistants}\label{sec:app_ai_usage}
We used AI for help rephrasing passages of this manuscript. 
All scientific claims, experimental design, results, and analysis are the authors' own, and we verified all AI-generated suggestions before incorporating them.

\section{Compute Requirements}\label{sec:compute_requirements}
For training the LUSR representations, we used four 40Gb A100s, each training run completing in under 2 hours.
We estimate that we ran roughly $100$ training runs, leaving us with $200$ hours of GPU time.
Training LUSR + DSVDD, we used roughly $2$ hours of compute time per run, training $5$ different models, thus totaling $10$ hours of compute time. 
All evaluations ran in a single 40Gb A100 GPU and used a negligible amount of compute compared to training.

\section{Hyper-parameters}
Training the LUSR representations, we used LoRA $r=32$ and $\alpha=64$, with a learning rate of $2e^{-5}$ over $3$ epochs.
Training LUSR + DSVDD, we trained for $50$ epochs (stopped at epoch $8$ with early stopping), using a batch size of $32$ per GPU (4 GPUs total), with a learning rate of $2e^{-5}$ and a learning rate warmup of $2000$ steps.

\end{document}